\documentclass[11pt]{article}

\usepackage[final]{acl}
\usepackage{booktabs}
\usepackage{multirow}
\usepackage{amsfonts}
\usepackage{times}
\usepackage{latexsym}
\usepackage{algorithm}
\usepackage{algorithmic}
\usepackage{amsmath}
\usepackage[T1]{fontenc}

\usepackage[utf8]{inputenc}

\usepackage{microtype}

\usepackage{inconsolata}

\usepackage{graphicx}

\usepackage[most]{tcolorbox}

\newtcolorbox{PromptBox}[2][]{%
    colback=gray!5!white,      
    colframe=gray!75!black,    
    title=\textbf{#2},         
    boxrule=0.8pt,             
    arc=3pt,                   
    left=6pt, right=6pt, top=6pt, bottom=6pt, 
    breakable,                 
    #1                         
}

\title{EviAgent: Evidence-Driven Agent for Radiology Report Generation}

\author{
  Tuoshi Qi\textsuperscript{1}\thanks{\ \ Equal contribution.} , 
  Shenshen Bu\textsuperscript{2}\footnotemark[1]\textsuperscript{,}\thanks{\ \ Corresponding authors.} , 
  Yingfei Xiang\textsuperscript{3} , 
  Zhiming Dai\textsuperscript{1}\footnotemark[2] \\
  \textsuperscript{1}School of Computer Science and Engineering, Sun Yat-sen University \\
  \textsuperscript{2}Shenzhen Institutes of Advanced Technology, Chinese Academy of Sciences \\
  \textsuperscript{3}Sangfor Technologies, Shenzhen, Guangdong, China \\
  \texttt{qitsh3@mail2.sysu.edu.cn}, \texttt{bushsh@alumni.sysu.edu.cn} \\
  \texttt{xiangyingfei@sangfor.com.cn}, \texttt{daizhim@mail.sysu.edu.cn}
}

\begin{document}
\pagestyle{plain}
\maketitle
\begin{abstract}
Automated radiology report generation holds immense potential to alleviate the heavy workload of radiologists. Despite the formidable vision-language capabilities of recent Multimodal Large Language Models (MLLMs), their clinical deployment is severely constrained by inherent limitations: their "black-box" decision-making renders the generated reports untraceable due to the lack of explicit visual evidence to support the diagnosis, and they struggle to access external domain knowledge. To address these challenges, we propose the Evidence-driven Radiology Report Generation Agent (EviAgent). Unlike opaque end-to-end paradigms, EviAgent coordinates a transparent reasoning trajectory by breaking down the complex generation process into granular operational units. We integrate multi-dimensional visual experts and retrieval mechanisms as external support modules, endowing the system with explicit visual evidence and high-quality clinical priors. Extensive experiments on MIMIC-CXR, CheXpert Plus, and IU-Xray datasets demonstrate that EviAgent outperforms both large-scale generalist models and specialized medical models, providing a robust and trustworthy solution for automated radiology report generation.
\end{abstract}

\section{Introduction}

Radiology reports play a crucial role in the medical diagnosis and treatment process. However, interpreting radiology images is highly time-consuming and dependent on expert experience, which has catalyzed research into automated radiology report generation. Early approaches based on Encoder-Decoder architectures \citep{chen-emnlp-2020-r2gen, endo2021retrieval, tanida2023interactive} made initial strides but were fundamentally limited by their reliance on statistical image-text correlations rather than region-specific visual evidence. Consequently, despite achieving high natural language generation metrics, these methods often suffer from severe factual errors in their generated clinical descriptions \citep{tanida2023interactive}.

Recent advancements have been primarily propelled by Multimodal Large Language Models (MLLMs). Generalist models \citep{openai2025gpt51, google2025gemini25, anthropic2025claude45s} have demonstrated formidable vision-language capabilities. Meanwhile, specialized open-source models like MedGemma \citep{sellergren2025medgemma} and Lingshu \citep{xu2025lingshu} have achieved performance comparable to generalist models with significantly fewer parameters (<10B) via high-quality biomedical instruction tuning. However, these end-to-end paradigms suffer from inherent limitations: First, they rely primarily on internal parametric reasoning, lacking the capability to acquire external knowledge; Second, their decision-making process is a "black box", rendering the generated reports untraceable due to the lack of explicit visual evidence to support the diagnosis. These issues severely constrain the performance of MLLMs in radiology report generation and hinder their potential for clinical deployment.

The emergence of agents in the medical domain \citep{kim2024mdagents, fallahpour2025medrax} has demonstrated that tool integration is an effective pathway to elevate model performance boundaries. However, most existing medical agents rely on close-source large models via cloud-based APIs as core planners. This data exfiltration paradigm inevitably raises severe concerns regarding medical data privacy and compliance.

To overcome these challenges, we propose the Evidence-driven Radiology Report Generation Agent (EviAgent) without additional training. We prioritize data privacy by building a fully local system where the core planner utilizes the open-source Qwen3-VL-8B \citep{bai2025qwen3vltechnicalreport} model and all integrated tools run entirely on-premise. Furthermore, by utilizing vLLM \citep{kwon2023efficientvllm}, a high-throughput and memory-efficient inference engine, our framework ensures strict privacy compliance and faster inference speeds, making the system both secure and efficient for clinical deployment.

We equip the planner with specialized local experts. We integrate discriminative perception tools to address the visual black box by explicitly localizing pathologies, ensuring diagnostic claims are physically anchored in the image. Furthermore, we incorporate a retrieval mechanism that functions analogously to a clinician's accumulated experience, enabling the system to acquire external knowledge beyond its internal parameters.

Our main contributions are as follows.

\begin{itemize}
\item We construct a dynamic multi-expert collaboration framework, breaking down the complex report generation process into a sequence of granular operational units. This design establishes a traceable decision trajectory that explicitly maps diagnostic findings back to their supporting visual evidence, thereby significantly enhancing the reliability of the final reports.

\item We integrate multi-dimensional visual expert models and knowledge bases as external support modules for evidence acquisition. This mechanism effectively overcomes the inherent limitation of MLLMs in retrieving external domain knowledge, endowing the system with explicit visual evidence and high-quality priors.

\item We conduct extensive experiments on the MIMIC-CXR \citep{johnson2019mimic}, CheXpert Plus \citep{chambon2024chexpert}, and IU-Xray \citep{iu_xray} datasets. The metrics demonstrate that our method outperforms both large-scale generalist models and specialized medical models, providing a robust and trustworthy solution for automated radiology report generation.
\end{itemize}

\section{Related Work}

Current advancements in automated radiology report generation are primarily categorized into two paradigms. The first involves Multimodal Large Language Models (MLLMs), which address radiological tasks via end-to-end foundation models. The second comprises medical agent systems, which enhance performance by either coordinating collaborative sub-agents or orchestrating specialized tools.

\subsection{MLLMs in Radiology}
Radiological image analysis encompasses a wide variety of clinical tasks, ranging from anatomical segmentation and lesion detection to Visual Question Answering (VQA) and report generation \citep{hosny2018artificialeee, najjar2023redefiningeee}. Recently, MLLMs have demonstrated significant potential in addressing these diverse challenges. Early works such as LLaVA-Med \citep{li2023llava} and Med-Flamingo \citep{moor2023med} explored aligning visual encoders with large language models to interpret radiological scans. More recently, advanced models like HuatuoGPT-V \citep{chen2024huatuogpt}, Lingshu \citep{xu2025lingshu}, and MedGemma \citep{sellergren2025medgemma} have further enhanced comprehensive performance across these visual understanding tasks through training on large-scale multimodal data. In parallel, specialized models \citep{chen2024chexagent, bannur2024maira2groundedradiologyreport, ZambranoChaves2025llavarad} have been developed specifically for report generation, which are explicitly trained on domain-specific image-report datasets to focus exclusively on this task. However, a critical limitation of these models is their inability to access external knowledge.

\subsection{Medical Agents}

Medical agent systems have emerged to surmount the rigidity of monolithic models by orchestrating specialized components. Contemporary research bifurcates into two streams. The first prioritizes collaborative reasoning, where frameworks like MedAgents \citep{wang2025medagent} and MDAgents \citep{kim2024mdagents} simulate medical consultations through role-playing and iterative debates to mitigate bias. The second stream focuses on multimodal tool integration. Systems such as MMedAgent \citep{li2024mmedagent} and MedAgent-Pro \citep{wang2025medagent} employ central planners to dispatch specialized tools, addressing various medical tasks.

\begin{figure*}[t]
  \centering
  \includegraphics[width=\textwidth]{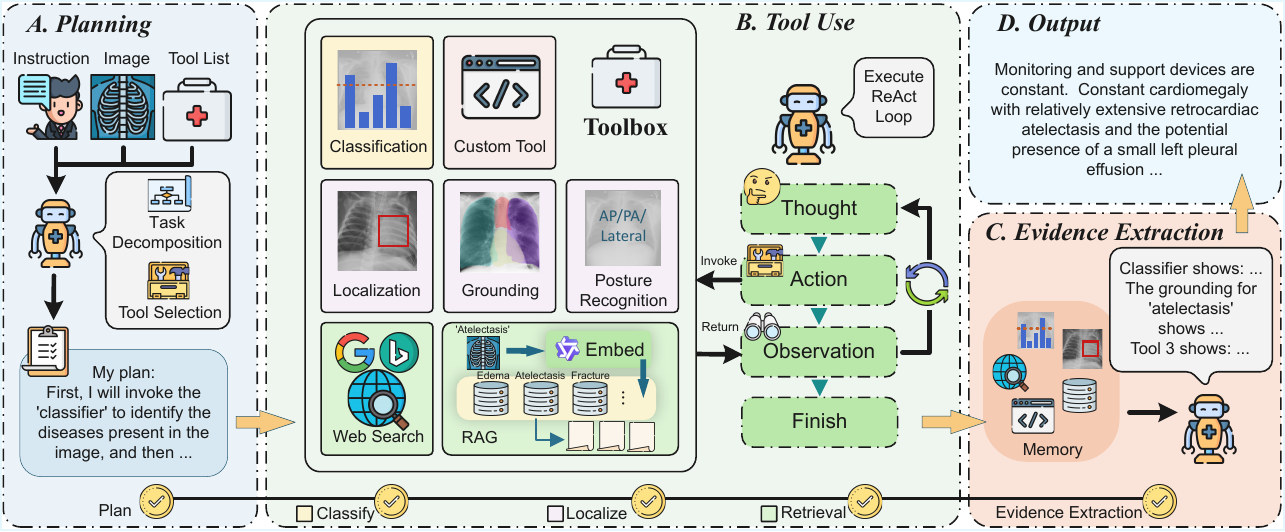}
  \caption{Overview of the EviAgent. The agent operates via four sequential stages: (A) \textit{Planning}, which decomposes the task into granular operational units; (B) \textit{Tool Use}, which dynamically invokes tools using the ReAct loop; (C) \textit{Evidence Extraction}, which consolidates discrete observations into traceable proofs; and (D) \textit{Output}, which synthesizes the final report strictly based on the accumulated evidence.}
  \label{fig:overall}
\end{figure*}

By leveraging collaboration or tool integration, these paradigms have pushed performance boundaries beyond MLLMs. However, they suffer from a persistent deficiency: a weak or absent connection between diagnostic conclusions and supporting visual evidence \citep{lou2025cxragent}. Meanwhile, they are not specifically designed for radiology report generation. Consequently, an evidence-driven radiology report generation agent that grounds diagnosis in explicit visual findings is urgently needed.

\section{The EviAgent}

We present EviAgent, an Evidence-driven Radiology Report Generation Agent. As illustrated in Figure \ref{fig:overall}, the agent coordinates the generation process through a structured agentic framework.

\subsection{The Evidence-based Framework}
The core of EviAgent is a progressive reasoning framework that transforms the "black-box" generation process into an evidence-based trajectory. Unlike end-to-end models that map pixels directly to text, our system coordinates the diagnosis through a structured "Plan-Act-Report" paradigm, where each diagnostic conclusion is rigorously derived from accumulated evidence.

\textbf{Planning and Decomposition.} Upon receiving a radiology image $I$ and a diagnostic instruction $Q$, the Planner $\mathcal{P}_{LLM}$ initiates the workflow by analyzing the global context. Instead of attempting to generate the report immediately, the model formulates a structured execution plan. The complex diagnostic goal is decomposed into a sequence of granular operational units (e.g., ``Detect lesions in the image'' and ``Localize diseases output by the classifier''). This explicit planning phase serves as a roadmap, ensuring that subsequent tool invocations are goal-oriented rather than random explorations.

\textbf{Tool-Augmented ReAct Loop.} Guided by the initial execution plan, the agent enters a ReAct \citep{yao2022react} loop. It is important to note that the plan generated in \textit{Planning} stage is inherently coarse-grained. It merely defines the high-level diagnostic scope but implies no knowledge of concrete pathological findings until the tools are actually executed. Therefore, this stage functions as a process of dynamic refinement. At each time step $t$, the Planner generates a thought trace based on the current state and dynamically selects the appropriate tool from the toolbox $\mathcal{T}$ to substantiate the preliminary plan with fine-grained evidence.

During execution, the agent parses the thought trace to determine the next action and its arguments. The selected tool is then executed, returning a structured result—such as a specific classification result or bounding box coordinates—which is immediately appended to the Evidence Memory $\mathcal{M}$. This continuous feedback loop empowers the agent with self-correction and investigative depth. For instance, if the initial screening tool detects atelectasis, the agent does not merely stop; instead, it spontaneously adapts its trajectory to trigger follow-up actions like ``localize the atelectasis'' thereby mimicking the iterative investigative process of a human radiologist.

\textbf{Evidence Extraction.} To ensure the integrity of the final report, the agent performs extraction, analysis, and integration of the accumulated tool outputs stored in Memory $\mathcal{M}$. This stage specifically targets valid execution results, including classification probabilities, localized visual features, and retrieved expert knowledge, extracting them from the verbose interaction history. By filtering out procedural artifacts and consolidating these findings, the system constructs a structured evidence chain $\mathcal{E}$. This chain functions as a purified set of clinical facts for the subsequent generation.

\textbf{Output.} Finally, the agent generates the final radiology report $R$ by strictly conditioning on the extracted evidence $\mathcal{E}$. By grounding the generation in these verifiable facts rather than purely parametric knowledge, EviAgent ensures that the output is not only linguistically coherent but also clinically accurate and evidence-based.

\subsection{Experts \& Knowledge Integration}

To support the tool-augmented ReAct loop, we construct a comprehensive toolbox tailored for radiological reasoning. By integrating a multi-dimensional set of specialized experts, we empower the Planner with professional visual perception and external domain knowledge. These experts are categorized into three distinct levels: Perception, Knowledge, and Customization.

\subsubsection{Discriminative Perception Experts}
This module functions as the visual perception unit, providing precise visual interpretation capabilities ranging from global screening to pixel-level grounding.

\textbf{Classification tool.}
For initial disease screening, we employ a Swin-Transformer \citep{bu2024ekagen} trained on MIMIC-CXR \citep{johnson2019mimic} as the core classifier. This model takes the raw chest X-ray image as input and performs multi-label classification to output a specific list of detected pathologies (e.g., "Pneumonia", "Cardiomegaly"). These classification results serve as the pivotal diagnostic anchors; the planner subsequently centers its entire reasoning trajectory, including localization and knowledge retrieval, specifically around these identified diseases.

\textbf{Localization tools.}
To enable spatial localization and detailed anatomical analysis, we integrate three specialized models for fine-grained perception:
\begin{itemize}
    \item Posture Recognition: Correctly identifying the projection view is a prerequisite for diagnosis. We employ CheXagent \citep{chen2024chexagent} for posture recognition.
    \item Visual Grounding: We utilize MAIRA-2 \citep{bannur2024maira2groundedradiologyreport} to handle grounding tasks. Given a specific disease query, it outputs precise bounding boxes (bbox), allowing the agent to spatially verify diagnoses.
    \item Anatomical Segmentation: For pixel-level precision, we incorporate MedSAM \citep{medsam} to segment anatomical structures and lesion boundaries.
\end{itemize}

\subsubsection{Retrieval-Augmented Knowledge}

Inspired by the clinical practice where radiologists rely on accumulated medical knowledge and experience to compose reports, we design this module to supplement the agent's generation process with external domain expertise. By retrieving historical cases relevant to the current visual findings, it provides the system with essential clinical priors.

To construct the external repository, we derive data from the MIMIC-CXR training set, partitioning it into $N=14$ pathology-specific knowledge bases corresponding to CheXpert \citep{irvin2019chexpert} standards. For each pathology $c$, we explicitly select $M=50$ high-quality triplet samples $(I_j, R_j, L_j)$, comprising the reference image, clinical report, and disease label respectively. We utilize the GME-Qwen2-VL \citep{zhang2024gme} model, denoted as $E_{mm}(\cdot)$, to compute multimodal embeddings. The offline embedding vector $\mathbf{v}_{c,j}$ is calculated as:
\begin{equation}
    \mathbf{v}_{c,j} = E_{mm}(I_j \oplus L_j)
\end{equation}
where $\oplus$ denotes concatenation. Consequently, each knowledge base $\mathcal{B}_c$ stores the paired mapping of embedding vectors and raw reports: $\{ (\mathbf{v}_{c,j}, R_j) \}_{j=1}^{M}$.

During online inference, the Planner analyzes the context and specifies a set of suspected pathology labels $\mathcal{C}_{query}$. Based on these labels, the RAG module routes the request to the corresponding knowledge bases. It generates queries for each target pathology and aggregates all candidates. The query generation, similarity scoring $s_{c,j}$, and final global retrieval $\mathcal{C}_{ret}$ are formulated as:
\begin{align}
    \mathbf{q}_c &= E_{mm}(I_{query} \oplus c), \quad \forall c \in \mathcal{C}_{query} \label{eq:query} \\
    s_{c,j} &= \frac{\mathbf{q}_c \cdot \mathbf{v}_{c,j}}{\|\mathbf{q}_c\| \|\mathbf{v}_{c,j}\|} \label{eq:sim} \\
    \mathcal{C}_{ret} &= \bigcup_{c \in \mathcal{C}_{query}} \{ R_j \mid \text{rank}(s_{c,j}) \leq k \} \label{eq:retrieval}
\end{align}
where $\text{rank}(\cdot)$ returns the descending rank of the similarity score among the candidate pool, ensuring only the top-$k$ most relevant reports are selected.

Additionally, we integrate a web search tool (via Bing or Google APIs). Serving as a dynamic knowledge extension, this module is designed to mitigate the inherent knowledge limitations of the foundation model, empowering the Planner to verify ambiguous concepts or acquire real-time medical information beyond the scope of its pre-trained parametric memory.

\subsubsection{Extensible Customization via Configuration}
We adopt the Model Context Protocol (MCP) as the unified interface layer, which provides a ``Plug-and-Play'' extension mechanism. Users can integrate local proprietary tools by simply modifying a JSON configuration file. By defining the tool's description, API endpoint, and argument schema, the Planner can automatically recognize and invoke these custom tools without any model fine-tuning. This design ensures that EviAgent can seamlessly integrate into diverse hospital information ecosystems, fostering a flexible and scalable deployment.

\begin{table*}[t]
    \centering
    \resizebox{\textwidth}{!}{
    \begin{tabular}{l|ccc|ccc|ccc}
        \toprule
        \multirow{2}{*}{\textbf{Model}} & \multicolumn{3}{c|}{\textbf{MIMIC-CXR}} & \multicolumn{3}{c|}{\textbf{CheXpert Plus}} & \multicolumn{3}{c}{\textbf{IU-Xray}} \\
        \cmidrule(lr){2-4} \cmidrule(lr){5-7} \cmidrule(lr){8-10}
        & RaTE & Semb & RadCliQ$^{-1}$ & RaTE & Semb & RadCliQ$^{-1}$ & RaTE & Semb & RadCliQ$^{-1}$ \\
        \midrule
        GPT-5.1 & 49.5 & 28.0 & 65.6 & \underline{48.0} & 27.5 & \textbf{49.2} & 56.8 & \underline{51.5} & 85.7 \\
        Claude 4.5 Sonnet & 49.1 & 23.9 & 64.7 & 47.4 & 22.5 & 48.1 & \underline{57.8} & 51.4 & 90.9 \\
        Gemini-2.5-Flash & 50.3 & 29.7 & 59.4 & 44.3 & 27.4 & 44.0 & 55.6 & 50.9 & 91.6 \\
        \midrule
        LLaVA-Med-7B & 12.8 & 18.3 & 52.9 & 38.8 & 23.5 & 44.0 & 40.9 & 16.0 & 58.1 \\
        HuatuoGPT-V-7B & 48.9 & 20.0 & 48.2 & 44.2 & 19.3 & 39.4 & 52.9 & 40.7 & 63.6 \\
        BiMediX2-8B & 44.4 & 17.7 & 53.0 & 40.8 & 21.6 & 43.3 & 40.1 & 11.6 & 53.8 \\
        MedGemma-4B-IT & \underline{52.4} & 29.2 & 62.9 & 47.2 & \underline{29.3} & 46.6 & 57.0 & 46.8 & 86.7 \\
        Lingshu-7B & 52.1 & \underline{30.0} & \underline{69.2} & 45.4 & 26.8 & 47.3 & 57.6 & 48.4 & \underline{108.1} \\
        \midrule
        InternVL2.5-8B & 47.0 & 21.0 & 56.2 & 43.1 & 19.7 & 42.7 & 51.1 & 36.7 & 67.0 \\
        InternVL3-8B & 48.2 & 21.5 & 55.1 & 44.3 & 25.2 & 43.7 & 51.2 & 31.3 & 59.9 \\
        Qwen2.5-VL-7B & 47.0 & 18.4 & 55.1 & 41.0 & 17.2 & 43.1 & 48.4 & 36.3 & 66.1 \\
        Qwen3-VL-8B & 48.9 & 26.1 & 64.2 & 45.9 & 27.3 & 44.6 & 50.3 & 46.5 & 74.7 \\
        \midrule
        \textbf{EviAgent (Ours)} & \textbf{52.6} & \textbf{43.6} & \textbf{76.6} & \textbf{49.8} & \textbf{30.4} & \underline{48.8} & \textbf{60.5} & \textbf{52.2} & \textbf{110.2} \\
        \bottomrule
    \end{tabular}
    }
    \caption{Results on automatic metrics. All scores are scaled by a factor 
of 100 to enhance clarity and comprehension. The best results are highlighted in \textbf{bold}, and the second-best results are \underline{underlined}. The data for medical MLLMs in this table are taken from the Lingshu paper \citep{xu2025lingshu}.}
\vspace{-1em}
\label{tab:main_results}
\end{table*}

\section{Experiments}

\subsection{Experimental Setup}

\noindent\textbf{Datasets.} We evaluate our framework on three public benchmarks. MIMIC-CXR \citep{johnson2019mimic}, a widely-used dataset developed by the Beth Israel Deaconess Medical Center. We utilize the official split, where the test set comprises 14 disease labels, 2,347 reports and 3,858 images. IU-Xray \citep{iu_xray}, a comprehensive chest X-ray dataset released by Indiana University. Following the established 7:1:2 partition from previous works \citep{wang2023metransformer, liu2021contrastive}, the test set includes 590 reports and 1,180 images. Additionally, we incorporate CheXpert Plus \citep{chambon2024chexpert}, a dataset containing 234 reports and 234 images in its official test split. For all three datasets, only the official test partitions are used in our experiments.

\noindent\textbf{Metrics.} Following prior studies \citep{bannur2024maira2groundedradiologyreport, zhang2024rexrank, xu2025lingshu}, we employ three advanced domain-specific metrics. RaTEScore \citep{zhao2024ratescore} is an entity-aware metric specifically designed to emphasize critical medical entities, including diagnostic findings and anatomical details. It demonstrates robustness to complex medical synonyms and high sensitivity to negation expressions. SembScore \citep{smit2020chexbertsemb} assesses clinical content alignment by calculating the cosine similarity between vectors of 14 pathological indicators, which are automatically extracted by the CheXbert labeler from both generated and ground-truth reports. Finally, we report RadCliQ$^{-1}$, the inverse of RadCliQ-v1 \citep{yu2023evaluatingradcliq}. This composite metric integrates BLEU for lexical precision, BERTScore for semantic consistency, and RadGraph-F1 for graph-based clinical relation matching, thereby providing a holistic assessment of report quality. We utilize the inverse form to ensure that higher scores consistently indicate better performance across all evaluations.

\noindent\textbf{Baselines.} To strictly evaluate the performance of our proposed framework, we compare EviAgent against a comprehensive set of state-of-the-art baselines categorized into three distinct groups:

(1) Close-Source Generalist MLLMs, representing the general reasoning capabilities, including GPT-5.1 \citep{openai2025gpt51}, Claude 4.5 Sonnet \citep{anthropic2025claude45s}, and Gemini-2.5-Flash \citep{google2025gemini25}.

(2) Medical Specialized MLLMs, which are explicitly optimized for the medical domain via continued pre-training or instruction tuning. Representative models include LLaVA-Med-7B \citep{li2023llava}, HuatuoGPT-V-7B \citep{chen2024huatuogpt}, BiMediX2-8B \citep{mullappilly2024bimedix2}, MedGemma-4B-IT \citep{sellergren2025medgemma}, and Lingshu-7B \citep{xu2025lingshu}.

(3) Open-Source Generalist MLLMs, serving as foundation baselines, including the InternVL series \citep{chen2024internvl25, zhu2025internvl3} and Qwen-VL series \citep{bai2025qwen2, bai2025qwen3vltechnicalreport}. 

\noindent\textbf{Settings.} We employ Qwen3-VL-8B-Instruct (Qwen3-VL-8B) \citep{bai2025qwen3vltechnicalreport} as the unified backbone, sequentially functioning as the planner, tool user, evidence extractor, and report generator. For the \textit{Tool Use} stage, the maximum number of interaction rounds is set to $T_{max}=10$ to prevent infinite loops while allowing sufficient exploration. In the RAG module, we retrieve the Top-$k=4$ reference reports. All experiments are run on the NVIDIA H100 GPU. The model is deployed via the vLLM library to accelerate inference.

\begin{table*}[t]
    \centering
    \resizebox{\textwidth}{!}{
    \begin{tabular}{l|cccc|cccc|cccc}
        \toprule
        \multirow{2}{*}{Model} & \multicolumn{4}{c|}{MIMIC-CXR} & \multicolumn{4}{c|}{CheXpert Plus} & \multicolumn{4}{c}{IU-Xray} \\
        \cmidrule(lr){2-5} \cmidrule(lr){6-9} \cmidrule(lr){10-13}
         & Acc & Loc & Prof & Adm & Acc & Loc & Prof & Adm & Acc & Loc & Prof & Adm \\
        \midrule
        GPT-5.1 & 4.91 & 4.98 & 8.67 & 5.62 & \underline{4.85} & 4.23 & 8.17 & \underline{5.55} & \underline{7.42} & 5.51 & 8.81 & 7.08 \\
        Claude 4.5 Sonnet & 3.51 & 3.79 & 8.59 & 4.40 & 3.41 & 3.14 & \underline{8.30} & 4.33 & 7.12 & 5.60 & \underline{9.15} & 7.20 \\
        Gemini-2.5-Flash & 5.74 & \underline{6.19} & \textbf{9.06} & \underline{6.48} & 4.28 & \underline{4.70} & 7.88 & 5.05 & 6.67 & 5.73 & 8.61 & 6.70 \\
        \midrule
        LLaVA-Med-7B & 1.74 & 2.88 & 4.28 & 2.23 & 1.71 & 2.85 & 3.86 & 2.06 & 2.07 & 3.37 & 3.86 & 2.01 \\
        HuatuoGPT-V-7B & 2.20 & 5.07 & 7.33 & 3.17 & 1.94 & 4.27 & 5.65 & 2.84 & 1.21 & 5.50 & 5.95 & 2.29 \\
        BiMediX2-8B & 1.41 & 2.76 & 3.84 & 1.86 & 1.22 & 2.35 & 3.69 & 1.68 & 0.51 & 2.77 & 4.27 & 1.14 \\
        MedGemma-4B-IT & 5.44 & 5.61 & 8.16 & 5.97 & 4.16 & 3.82 & 7.68 & 4.80 & 7.24 & 6.28 & 8.38 & 7.41 \\
        Lingshu-7B & \underline{5.88} & 5.87 & 8.66 & 6.37 & 4.60 & 3.91 & 7.75 & 5.20 & 7.39 & 6.24 & 8.88 & \underline{7.67} \\
        \midrule
        InternVL2.5-8B & 2.41 & 3.44 & 7.33 & 3.41 & 2.55 & 2.86 & 7.06 & 3.52 & 6.81 & 5.87 & 8.00 & 6.58 \\
        InternVL3-8B & 3.07 & 4.71 & 7.55 & 4.05 & 2.82 & 3.88 & 7.29 & 3.85 & 3.68 & \underline{6.36} & 7.72 & 4.48 \\
        Qwen2.5-VL-7B & 2.30 & 3.49 & 7.30 & 3.43 & 2.21 & 2.93 & 7.13 & 3.33 & 5.79 & 6.04 & 7.86 & 5.91 \\
        Qwen3-VL-8B & 3.94 & 4.66 & 7.51 & 4.75 & 3.72 & 3.78 & 7.48 & 4.58 & 5.74 & 4.98 & 7.63 & 5.64 \\
        \midrule
        \textbf{EviAgent (Ours)} & \textbf{6.04} & \textbf{6.32} & \underline{8.70} & \textbf{6.61} & \textbf{4.91} & \textbf{6.66} & \textbf{8.45} & \textbf{5.72} & \textbf{7.48} & \textbf{6.55} & \textbf{9.29} & \textbf{7.72} \\
        \bottomrule
    \end{tabular}
    }
    \caption{Clinical value evaluation via LLM-as-a-Judge. We evaluate the generated reports across four dimensions: Acc (Accuracy), Loc (Localization), Prof (Professionalism), and Adm (Clinical Admissibility).}
    \label{tab:llm_judge}
\end{table*}

\subsection{Performance on Multiple Metrics}
Table \ref{tab:main_results} demonstrates that EviAgent achieves the best performance across three datasets on almost all metrics. On the large-scale MIMIC-CXR dataset, our method establishes a significant lead, surpassing close-source giants like GPT-5.1 by 3.1 in RaTE and outperforming specialized medical MLLMs such as Lingshu-7B by 13.6 in Semb. Furthermore, on the IU-Xray and CheXpert Plus benchmarks, EviAgent consistently exceeds the baselines, achieving a RadCliQ$^{-1}$ of 110.2 on IU-Xray. This observation indicates that the proposed framework achieves better performance compared to standard MLLMs.

The efficacy of our framework is most evident when comparing EviAgent with its backbone, Qwen3-VL-8B. Across all three datasets, equipping the base model with our agentic workflow yields a substantial performance leap. For instance, Semb scores on MIMIC-CXR nearly double from 26.1 to 43.6, and RaTE on IU-Xray improves by 10.2 points. GPT-5.1 shows slight linguistic advantages on the RadCliQ$^{-1}$ metric for CheXpert Plus due to extensive general pre-training. These results validate the effectiveness of our proposed method, demonstrating that the collaborative synergy of multiple tools within the agentic architecture enables the system to significantly outperform the traditional standalone inference baseline.

\subsection{Clinical Value Evaluation via LLM-as-a-Judge}
To further assess the granular medical details and holistic clinical validity required in real-world diagnosis, we conducted a comprehensive evaluation on the entire test set of each benchmark using DeepSeek-V3.2 \citep{deepseekai2025deepseekv32pushingfrontieropen} as an impartial judge. The judge scored the generated reports on a scale of 0 to 10 across four specific dimensions: Accuracy, which measures the correctness of findings; Localization, evaluating the precision of anatomical descriptions; Professionalism, assessing the use of standardized radiological terminology; and Clinical Admissibility, a holistic metric indicating whether the report is robust enough for clinical workflows without major modification. As shown in Table \ref{tab:llm_judge}, EviAgent achieves the best performance across almost all metrics. While Gemini-2.5-Flash exhibits a marginal lead in Professionalism for MIMIC-CXR due to the inherent linguistic advantages of large-scale models, our framework consistently dominates in all other critical diagnostic dimensions.

\begin{table}[t]
    \centering
    
    \renewcommand{\arraystretch}{1.15}
    \setlength{\tabcolsep}{3pt}
    \resizebox{\columnwidth}{!}
    {
    \begin{tabular}{c c c c c c c} 
        \toprule
        \multirow{2}{*}{Dataset} & \multirow{2}{*}{Metric} & \multirow{2}{*}{Full} & \multicolumn{3}{c}{w/o (Remove)} & \multirow{2}{*}{Base} \\
        
        \cmidrule(lr){4-6} 
        
         & &  & Cls. & Loc. & Ret. &  \\
        \midrule
        
        \multirow{3}{*}{\rotatebox[origin=c]{90}{\shortstack{MIMIC\\-CXR}}} 
         & RaTE & \textbf{52.6} & 50.6 & 51.8 & 51.7 & 48.9 \\
         & Semb & \textbf{43.6} & 23.1 & 43.0 & 43.4 & 26.1 \\
         & RadCliQ$^{-1}$ & \textbf{76.6} & 65.1 & 75.3 & 76.2 & 64.2 \\
        \midrule
        
        \multirow{3}{*}{\rotatebox[origin=c]{90}{\shortstack{CheXpert\\Plus}}} 
         & RaTE & \textbf{49.8} & 47.7 & 49.2 & 48.5 & 45.9 \\
         & Semb & \textbf{30.4} & 27.5 & 29.9 & 28.4 & 27.3 \\
         & RadCliQ$^{-1}$ & \textbf{48.8} & 47.4 & 48.2 & 47.9 & 44.6 \\
        \midrule
        
        \multirow{3}{*}{\rotatebox[origin=c]{90}{\shortstack{IU\\-Xray}}} 
         & RaTE & \textbf{60.5} & 54.7 & 60.1 & 59.0 & 50.3 \\
         & Semb & \textbf{52.2} & 43.1 & 51.5 & 50.4 & 46.5 \\
         & RadCliQ$^{-1}$ & \textbf{110.2} & 83.9 & 104.6 & 91.8 & 74.7 \\
         
        \bottomrule
    \end{tabular}
    }
    \caption{Ablation study of module contributions. We report the performance drop when removing specific components. Full represents the complete EviAgent framework, while Base denotes the vanilla Qwen3-VL-8B performing end-to-end generation. Cls.: Classification, Loc.: Localization, Ret.: Retrieval.}
    \label{tab:ablation}
\end{table}

\begin{table}[t]
    \centering
    \renewcommand{\arraystretch}{1.3}
    \resizebox{\columnwidth}{!}{
    \begin{tabular}{@{} l c c c @{}} 
        \toprule
         & \textbf{MIMIC-CXR} & \textbf{CheXpert Plus} & \textbf{IU-Xray} \\
        \midrule
        EviAgent & \textbf{76.6} & \textbf{48.8} & \textbf{110.2} \\
        w/o Planning & 70.4 & 44.7 & 101.7 \\
        w/o Evidence Extraction & 75.8 & 47.1 & 106.8 \\
        \bottomrule
    \end{tabular}
    }
    \caption{Ablation study on the core reasoning mechanisms of EviAgent. The reported metric is RadCliQ$^{-1}$. Removing the planning or evidence extraction stages leads to a degradation in report quality.}
    \label{tab:ablation_mechanisms}
\end{table}

\subsection{Ablation Study}
To validate individual module contributions, we evaluated variants by removing the \textbf{Classification}, \textbf{Localization}, and \textbf{Retrieval} components (denoted as ``w/o''). Table \ref{tab:ablation} summarizes the results.

Removing the classification expert causes the most severe degradation. Notably, the Semb score on MIMIC-CXR drops to 23.1, even lower than the vanilla backbone. We attribute this to error propagation: without reliable triage, the planner's initial hallucinations misguide subsequent tool invocations, causing the reasoning trajectory to deviate progressively further than standard end-to-end generation.

The absence of localization tools leads to a decline in RaTE across all datasets, confirming that visual grounding is essential for verifying fine-grained anatomical entities. Similarly, removing the retrieval tools consistently lowers RadCliQ$^{-1}$, indicating that external knowledge is vital for maintaining professional stylistic standards and minimizing linguistic hallucinations.

Furthermore, we investigate the impact of the core reasoning mechanisms: Planning and Evidence Extraction. As shown in Table \ref{tab:ablation_mechanisms}, removing the planning stage results in a significant performance drop across all three datasets (e.g., from 110.2 to 101.7 on IU-Xray). This highlights the necessity of structured, step-by-step reasoning before tool invocation, without which the agent struggles to formulate a coherent diagnostic strategy. Similarly, omitting the evidence extraction module leads to a consistent decrease in performance. This demonstrates that explicitly filtering and summarizing tool outputs helps in reducing noise and extracting salient clinical findings, ultimately generating more accurate final reports.

\begin{figure*}[t]
  \centering
  \includegraphics[width=\textwidth]{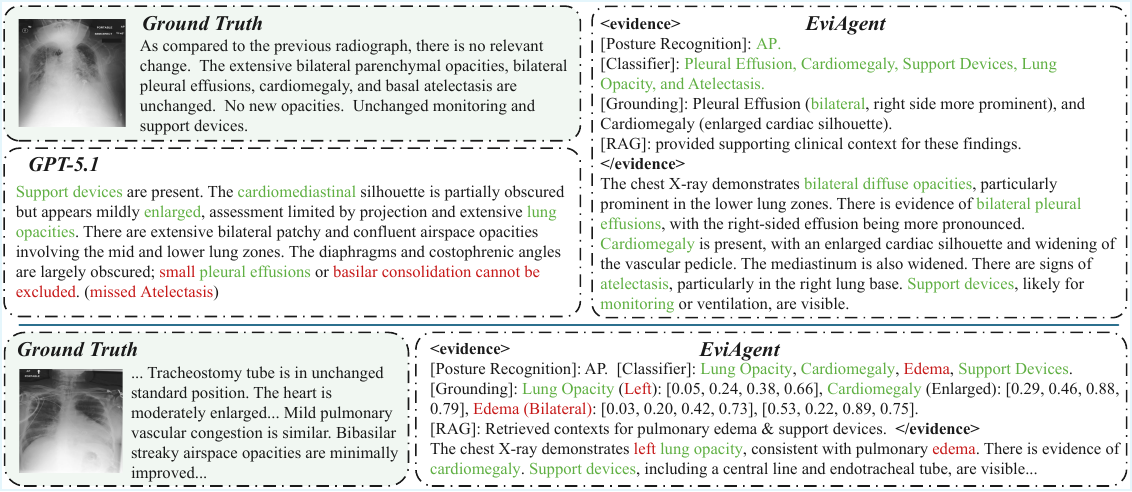}
  \caption{Qualitative analysis. Top: A complex case (Study 56122911) where EviAgent achieves accurate diagnosis and localization. Bottom: A case demonstrating error traceability (Study 50239281). Text in green indicates consistency with the ground truth, while text in red denotes discrepancies.}
  \label{fig:qualitative_cases}
\end{figure*}

\subsection{Qualitative Analysis}
To qualitatively validate the efficacy and interpretability of our framework, we present a comparative analysis of two cases in Figure \ref{fig:qualitative_cases}.

\textbf{Precision in Complex Clinical Scenarios.}
The top panel illustrates a complex ICU scenario. The ground truth confirms extensive bilateral opacities, bilateral pleural effusions, cardiomegaly, and basal atelectasis. In this case, the generalist model GPT-5.1 exhibits significant failures: it misses the diagnosis of atelectasis and incorrectly characterizes the effusions as small. Furthermore, it fails to provide a definitive conclusion, stating that effusions or consolidation cannot be excluded. In contrast, EviAgent successfully identifies all key pathologies. The tools correctly capture the bilateral nature of the effusions and opacities, as well as the presence of cardiomegaly and atelectasis. This structured evidence guides the planner to generate a precise report that accurately reflects the extent and spatial distribution of the disease.

\textbf{Traceability of Diagnostic Discrepancies.}
The bottom panel demonstrates the error traceability of our architecture. In this case, the ground truth describes bibasilar opacities and mild congestion. However, our agent reports a left-sided opacity and pulmonary edema. Crucially, this discrepancy is not a generative fabrication. The evidence log explicitly shows that the grounding tool restricted the lung opacity to the left field and the classifier outputted edema. The planner faithfully aggregated these specific tool outputs into the final report. This confirms that the errors—mislocalization and clinical over-diagnosis—stem directly from the perception module's limitations rather than the reasoning engine. Crucially, this validates the integrity of our evidence-driven paradigm: the planner faithfully executes based on the provided visual signals rather than hallucinating content. Consequently, the system maintains full traceability, allowing diagnostic discrepancies to be logically isolated to specific visual experts.

\begin{table}[t]
    \centering
    \resizebox{\columnwidth}{!}
    {
    \begin{tabular}{lcccc}
        \toprule
        Backbone Planner & VR & Tool Call & FER & QS \\
        \midrule
        GPT-5.1 & 98.9\% & 4.40 & 4.2\% & 6.95 \\
        Claude 4.5 Sonnet & 98.6\% & 4.48 & 4.5\% & 6.93 \\
        InternVL3-8B & 97.2\% & 5.13 & 10.3\% & 6.72 \\
        Qwen3-VL-8B & 98.2\% & 4.70 & 8.8\% & 6.89 \\
        \bottomrule
    \end{tabular}
    }
    \vspace{0.2cm}
    \caption{Scalability analysis of different backbone planners on the mixed evaluation subset.}
    \label{tab:agent_scalability}
\end{table}

\subsection{Agentic Capability and Extensibility Analysis}
To evaluate the extensibility of our framework and the impact of different backbone planners, we conducted a test using a mixed evaluation subset of 1,000 instances: 500 cases from MIMIC-CXR to test report generation capabilities and 500 questions from MIMIC-CXR-VQA \citep{bae2023ehrxqamultimodalquestionanswering}.

We employ four agent-specific metrics: (1) Valid Rate (VR): The probability of generating a valid final response within the maximum limit of tool invocations $T_{max}$; (2) Tool Call: The average number of tool executions per episode; (3) Format Error Rate (FER): The frequency of parsing failures or hallucinating tools; and (4) Quality Score (QS): The generation quality ranging from 0 to 10 by DeepSeek-V3.2 based on the ground truth.

As shown in Table \ref{tab:agent_scalability}, although Qwen3-VL-8B incurs higher average tool calls than GPT-5.1 due to retry loops triggered by formatting errors, its final Quality Score is only marginally lower (6.89 vs. 6.95). Our experiment consistently demonstrate the stability and robustness of the proposed agent framework regardless of the underlying model choice.

\section{Conclusion}
In this work, we presented EviAgent, an evidence-driven agent designed to fundamentally address the twin challenges of ``black-box'' opacity and the inability to access external knowledge in existing MLLMs. By shifting from static parametric reasoning to active evidence acquisition, our system ensures that every diagnostic conclusion is rigorously grounded in explicit visual findings and retrieved domain knowledge. Extensive experiments demonstrate that EviAgent achieves superior performance in clinical accuracy. Notably, our approach significantly outperforms open-source models of similar scale and even surpasses top-tier proprietary giants like GPT-5.1, validating the efficacy of our evidence-driven agentic framework in medical domains.

\section*{Limitations}

While EviAgent demonstrates superior clinical accuracy and interpretability, it entails a trade-off regarding inference latency. Unlike monolithic end-to-end MLLMs that generate reports in a single forward pass, our framework relies on a multi-round evidence-driven agentic workflow. The frequent invocation of external tools and the iterative reasoning process inevitably introduce additional computational overhead, resulting in longer generation times compared to direct generation approaches.

Future optimizations in tool execution and infrastructure are expected to mitigate this latency, bridging the efficiency gap without compromising clinical rigor.

\section*{Ethical Considerations}

This research is conducted using the MIMIC-CXR \citep{johnson2019mimic}, MIMIC-CXR-VQA \citep{bae2023ehrxqamultimodalquestionanswering}, IU-Xray \citep{iu_xray}, and CheXpert Plus \citep{chambon2024chexpert} datasets, all of which are publicly accessible and have undergone automatic de-identification to mitigate privacy risks.

While our framework ensures that the generated findings are explicitly traceable to specific tool outputs, we emphasize that such traceability does not guarantee absolute accuracy. The integrated tools or the reasoning process may still yield erroneous evidence or omissions. Therefore, these outputs must not be used as a substitute for expert medical judgment. We strongly advocate for mandatory validation by qualified radiologists or healthcare professionals before any clinical or diagnostic application.


\bibliography{main}

\begin{thebibliography}{43}
\providecommand{\natexlab}[1]{#1}

\bibitem[{{Anthropic}(2025)}]{anthropic2025claude45s}
{Anthropic}. 2025.
\newblock \href {https://www.anthropic.com/news/claude-sonnet-4-5} {Introducing {Claude} {Sonnet} 4.5}.
\newblock Accessed: 2025-12-27.

\bibitem[{Bae et~al.(2023)Bae, Kyung, Ryu, Cho, Lee, Kweon, Oh, Ji, Chang, Kim, and Choi}]{bae2023ehrxqamultimodalquestionanswering}
Seongsu Bae, Daeun Kyung, Jaehee Ryu, Eunbyeol Cho, Gyubok Lee, Sunjun Kweon, Jungwoo Oh, Lei Ji, Eric I-Chao Chang, Tackeun Kim, and Edward Choi. 2023.
\newblock \href {https://arxiv.org/abs/2310.18652} {Ehrxqa: A multi-modal question answering dataset for electronic health records with chest x-ray images}.
\newblock \emph{Preprint}, arXiv:2310.18652.

\bibitem[{Bai et~al.(2025{\natexlab{a}})Bai, Cai, Chen, Chen, Chen, Cheng, Deng, Ding, Gao, Ge, Ge, Guo, Huang, Huang, Huang, Hui, Jiang, Li, Li, Li, Li, Lin, Lin, Liu, Liu, Liu, Liu, Liu, Liu, Lu, Luo, Lv, Men, Meng, Ren, Ren, Song, Sun, Tang, Tu, Wan, Wang, Wang, Wang, Wang, Xie, Xu, Xu, Xu, Yang, Yang, Yang, Yang, Yu, Zhang, Zhang, Zhang, Zheng, Zhong, Zhou, Zhou, Zhou, Zhu, and Zhu}]{bai2025qwen3vltechnicalreport}
Shuai Bai, Yuxuan Cai, Ruizhe Chen, Keqin Chen, Xionghui Chen, Zesen Cheng, Lianghao Deng, Wei Ding, Chang Gao, Chunjiang Ge, Wenbin Ge, Zhifang Guo, Qidong Huang, Jie Huang, Fei Huang, Binyuan Hui, Shutong Jiang, Zhaohai Li, Mingsheng Li, and 45 others. 2025{\natexlab{a}}.
\newblock \href {https://arxiv.org/abs/2511.21631} {Qwen3-vl technical report}.
\newblock \emph{Preprint}, arXiv:2511.21631.

\bibitem[{Bai et~al.(2025{\natexlab{b}})Bai, Chen, Liu, Wang, Ge, Song, Dang, Wang, Wang, Tang et~al.}]{bai2025qwen2}
Shuai Bai, Keqin Chen, Xuejing Liu, Jialin Wang, Wenbin Ge, Sibo Song, Kai Dang, Peng Wang, Shijie Wang, Jun Tang, and 1 others. 2025{\natexlab{b}}.
\newblock Qwen2. 5-vl technical report.
\newblock \emph{arXiv preprint arXiv:2502.13923}.

\bibitem[{Bannur et~al.(2024)Bannur, Bouzid, Castro, Schwaighofer, Thieme, Bond-Taylor, Ilse, Pérez-García, Salvatelli, Sharma, Meissen, Ranjit, Srivastav, Gong, Codella, Falck, Oktay, Lungren, Wetscherek, Alvarez-Valle, and Hyland}]{bannur2024maira2groundedradiologyreport}
Shruthi Bannur, Kenza Bouzid, Daniel~C. Castro, Anton Schwaighofer, Anja Thieme, Sam Bond-Taylor, Maximilian Ilse, Fernando Pérez-García, Valentina Salvatelli, Harshita Sharma, Felix Meissen, Mercy Ranjit, Shaury Srivastav, Julia Gong, Noel C.~F. Codella, Fabian Falck, Ozan Oktay, Matthew~P. Lungren, Maria~Teodora Wetscherek, and 2 others. 2024.
\newblock \href {https://arxiv.org/abs/2406.04449} {Maira-2: Grounded radiology report generation}.
\newblock \emph{Preprint}, arXiv:2406.04449.

\bibitem[{Bu et~al.(2024)Bu, Li, Yang, and Dai}]{bu2024ekagen}
Shenshen Bu, Taiji Li, Yuedong Yang, and Zhiming Dai. 2024.
\newblock Instance-level expert knowledge and aggregate discriminative attention for radiology report generation.
\newblock In \emph{Proceedings of the IEEE/CVF Conference on Computer Vision and Pattern Recognition}, pages 14194--14204.

\bibitem[{Chambon et~al.(2024)Chambon, Delbrouck, Sounack, Huang, Chen, Varma, Truong, Chuong, and Langlotz}]{chambon2024chexpert}
Pierre Chambon, Jean-Benoit Delbrouck, Thomas Sounack, Shih-Cheng Huang, Zhihong Chen, Maya Varma, Steven~QH Truong, Chu~The Chuong, and Curtis~P Langlotz. 2024.
\newblock Chexpert plus: Augmenting a large chest x-ray dataset with text radiology reports, patient demographics and additional image formats.
\newblock \emph{arXiv preprint arXiv:2405.19538}.

\bibitem[{Chen et~al.(2024{\natexlab{a}})Chen, Gui, Ouyang, Gao, Chen, Chen, Wang, Zhang, Cai, Ji et~al.}]{chen2024huatuogpt}
Junying Chen, Chi Gui, Ruyi Ouyang, Anningzhe Gao, Shunian Chen, Guiming~Hardy Chen, Xidong Wang, Ruifei Zhang, Zhenyang Cai, Ke~Ji, and 1 others. 2024{\natexlab{a}}.
\newblock Huatuogpt-vision, towards injecting medical visual knowledge into multimodal llms at scale.
\newblock \emph{arXiv preprint arXiv:2406.19280}.

\bibitem[{Chen et~al.(2024{\natexlab{b}})Chen, Wang, Cao, Liu, Gao, Cui, Zhu, Ye, Tian, Liu et~al.}]{chen2024internvl25}
Zhe Chen, Weiyun Wang, Yue Cao, Yangzhou Liu, Zhangwei Gao, Erfei Cui, Jinguo Zhu, Shenglong Ye, Hao Tian, Zhaoyang Liu, and 1 others. 2024{\natexlab{b}}.
\newblock Expanding performance boundaries of open-source multimodal models with model, data, and test-time scaling.
\newblock \emph{arXiv preprint arXiv:2412.05271}.

\bibitem[{Chen et~al.(2020)Chen, Song, Chang, and Wan}]{chen-emnlp-2020-r2gen}
Zhihong Chen, Yan Song, Tsung-Hui Chang, and Xiang Wan. 2020.
\newblock Generating radiology reports via memory-driven transformer.
\newblock In \emph{Proceedings of the 2020 Conference on Empirical Methods in Natural Language Processing}.

\bibitem[{Chen et~al.(2024{\natexlab{c}})Chen, Varma, Delbrouck, Paschali, Blankemeier, Van~Veen, Valanarasu, Youssef, Cohen, Reis et~al.}]{chen2024chexagent}
Zhihong Chen, Maya Varma, Jean-Benoit Delbrouck, Magdalini Paschali, Louis Blankemeier, Dave Van~Veen, Jeya Maria~Jose Valanarasu, Alaa Youssef, Joseph~Paul Cohen, Eduardo~Pontes Reis, and 1 others. 2024{\natexlab{c}}.
\newblock Chexagent: Towards a foundation model for chest x-ray interpretation.
\newblock In \emph{AAAI 2024 Spring Symposium on Clinical Foundation Models}.

\bibitem[{DeepSeek-AI et~al.(2025)DeepSeek-AI, Liu, Mei, Lin, Xue, Wang, Xu, Wu, Zhang, Lin, Dong, Lu, Zhao, Deng, Xu, Ruan, Dai, Guo, Yang, Chen, Li, Zhou, Lin, Dai, Hao, Chen, Li, Zhang, Xu, Li, Liang, Wei, Zhang, Luo, Ji, Ding, Tang, Cao, Gao, Qu, Zeng, Huang, Li, Xu, Hu, Chen, Xiang, Yuan, Cheng, Zhu, Ran, Jiang, Qiu, Li, Song, Dong, Gao, Guan, Huang, Zhou, Huang, Yu, Wang, Zhang, Wang, Zhao, Yin, Guo, Luo, Ma, Wang, Zhang, Di, Xu, Zhang, Zhang, Tang, Zhou, Huang, Cong, Wang, Wang, Zhu, Li, Chen, Du, Xu, Ge, Zhang, Pan, Wang, Yin, Xu, Shen, Zhang, Liu, Lu, Zhou, Chen, Cai, Chen, Hu, Liu, Hu, Ma, Wang, Yu, Zhou, Pan, Zhou, Ni, Yun, Pei, Ye, Yue, Zeng, Liu, Liang, Pang, Luo, Gao, Zhang, Gao, Wang, Bi, Liu, Wang, Chen, Zhang, Nie, Cheng, Liu, Xie, Liu, Yu, Li, Yang, Li, Chen, Su, Pan, Lin, Fu, Wang, Zhang, Xu, Ma, Li, Li, Zhao, Sun, Wang, Qian, Yu, Zhang, Ding, Shi, Xiong, He, Zhou, Zhong, Piao, Wang, Chen, Tan, Wei, Ma, Liu, Yang, Guo, Wu, Wu, Cheng, Ou, Xu, Wang, Gong, Wu, Zou, Li, Xiong, Luo, You, Liu,
  Zhou, Wu, Ren, Zhao, Ren, Sha, Fu, Xu, Xie, Zhang, Hao, Gou, Ma, Yan, Shao, Huang, Wu, Li, Zhang, Xu, Wang, Gu, Zhu, Li, Zhang, Xie, Gao, Pan, Yao, Feng, Li, Cai, Ni, Xu, Li, Tian, Chen, Jin, Li, Zhou, Sun, Li, Jin, Shen, Chen, Song, Zhou, Zhu, Huang, Li, Zheng, Zhu, Ma, Huang, Xu, Zhang, Ji, Liang, Guo, Chen, Xia, Wang, Li, Zhang, Chen, Sun, Wu, Ye, Wang, Xiao, An, Wang, Sun, Wang, Tang, Zha, Zhang, Ju, Zhang, and Qu}]{deepseekai2025deepseekv32pushingfrontieropen}
DeepSeek-AI, Aixin Liu, Aoxue Mei, Bangcai Lin, Bing Xue, Bingxuan Wang, Bingzheng Xu, Bochao Wu, Bowei Zhang, Chaofan Lin, Chen Dong, Chengda Lu, Chenggang Zhao, Chengqi Deng, Chenhao Xu, Chong Ruan, Damai Dai, Daya Guo, Dejian Yang, and 245 others. 2025.
\newblock \href {https://arxiv.org/abs/2512.02556} {Deepseek-v3.2: Pushing the frontier of open large language models}.
\newblock \emph{Preprint}, arXiv:2512.02556.

\bibitem[{Demner-Fushman et~al.(2015)Demner-Fushman, Kohli, Rosenman, Shooshan, Rodriguez, Antani, Thoma, and McDonald}]{iu_xray}
Dina Demner-Fushman, Marc~D Kohli, Marc~B Rosenman, Sonya~E Shooshan, Laritza Rodriguez, Sameer Antani, George~R Thoma, and Clement~J McDonald. 2015.
\newblock Preparing a collection of radiology examinations for distribution and retrieval.
\newblock \emph{Journal of the American Medical Informatics Association}, 23(2):304--310.

\bibitem[{Endo et~al.(2021)Endo, Krishnan, Krishna, Ng, and Rajpurkar}]{endo2021retrieval}
Mark Endo, Rayan Krishnan, Viswesh Krishna, Andrew~Y Ng, and Pranav Rajpurkar. 2021.
\newblock Retrieval-based chest x-ray report generation using a pre-trained contrastive language-image model.
\newblock In \emph{Machine Learning for Health}, pages 209--219. PMLR.

\bibitem[{Fallahpour et~al.(2025)Fallahpour, Ma, Munim, Lyu, and Wang}]{fallahpour2025medrax}
Adibvafa Fallahpour, Jun Ma, Alif Munim, Hongwei Lyu, and Bo~Wang. 2025.
\newblock Medrax: Medical reasoning agent for chest x-ray.
\newblock \emph{arXiv preprint arXiv:2502.02673}.

\bibitem[{{Google DeepMind}(2025)}]{google2025gemini25}
{Google DeepMind}. 2025.
\newblock \href {https://blog.google/technology/google-deepmind/gemini-model-thinking-updates-march-2025/#gemini-2-5-thinking} {Gemini 2.5: Our most intelligent ai model}.
\newblock Accessed: 2025-12-27.

\bibitem[{Hosny et~al.(2018)Hosny, Parmar, Quackenbush, Schwartz, and Aerts}]{hosny2018artificialeee}
Ahmed Hosny, Chintan Parmar, John Quackenbush, Lawrence~H Schwartz, and Hugo~JWL Aerts. 2018.
\newblock Artificial intelligence in radiology.
\newblock \emph{Nature Reviews Cancer}, 18(8):500--510.

\bibitem[{Irvin et~al.(2019)Irvin, Rajpurkar, Ko, Yu, Ciurea-Ilcus, Chute, Marklund, Haghgoo, Ball, Shpanskaya et~al.}]{irvin2019chexpert}
Jeremy Irvin, Pranav Rajpurkar, Michael Ko, Yifan Yu, Silviana Ciurea-Ilcus, Chris Chute, Henrik Marklund, Behzad Haghgoo, Robyn Ball, Katie Shpanskaya, and 1 others. 2019.
\newblock Chexpert: A large chest radiograph dataset with uncertainty labels and expert comparison.
\newblock In \emph{Proceedings of the AAAI conference on artificial intelligence}, volume~33, pages 590--597.

\bibitem[{Johnson et~al.(2019)Johnson, Pollard, Berkowitz, Greenbaum, Lungren, Deng, Mark, and Horng}]{johnson2019mimic}
Alistair~EW Johnson, Tom~J Pollard, Seth~J Berkowitz, Nathaniel~R Greenbaum, Matthew~P Lungren, Chih-ying Deng, Roger~G Mark, and Steven Horng. 2019.
\newblock Mimic-cxr, a de-identified publicly available database of chest radiographs with free-text reports.
\newblock \emph{Scientific data}, 6(1):317.

\bibitem[{Kim et~al.(2024)Kim, Park, Jeong, Chan, Xu, McDuff, Lee, Ghassemi, Breazeal, and Park}]{kim2024mdagents}
Yubin Kim, Chanwoo Park, Hyewon Jeong, Yik~S Chan, Xuhai Xu, Daniel McDuff, Hyeonhoon Lee, Marzyeh Ghassemi, Cynthia Breazeal, and Hae~W Park. 2024.
\newblock Mdagents: An adaptive collaboration of llms for medical decision-making.
\newblock \emph{Advances in Neural Information Processing Systems}, 37:79410--79452.

\bibitem[{Kwon et~al.(2023)Kwon, Li, Zhuang, Sheng, Zheng, Yu, Gonzalez, Zhang, and Stoica}]{kwon2023efficientvllm}
Woosuk Kwon, Zhuohan Li, Siyuan Zhuang, Ying Sheng, Lianmin Zheng, Cody~Hao Yu, Joseph Gonzalez, Hao Zhang, and Ion Stoica. 2023.
\newblock Efficient memory management for large language model serving with pagedattention.
\newblock In \emph{Proceedings of the 29th symposium on operating systems principles}, pages 611--626.

\bibitem[{Li et~al.(2024)Li, Yan, Pan, Luo, Ji, Ding, Xu, Liu, Dong, Lin et~al.}]{li2024mmedagent}
Binxu Li, Tiankai Yan, Yuanting Pan, Jie Luo, Ruiyang Ji, Jiayuan Ding, Zhe Xu, Shilong Liu, Haoyu Dong, Zihao Lin, and 1 others. 2024.
\newblock Mmedagent: Learning to use medical tools with multi-modal agent.
\newblock \emph{arXiv preprint arXiv:2407.02483}.

\bibitem[{Li et~al.(2023)Li, Wong, Zhang, Usuyama, Liu, Yang, Naumann, Poon, and Gao}]{li2023llava}
Chunyuan Li, Cliff Wong, Sheng Zhang, Naoto Usuyama, Haotian Liu, Jianwei Yang, Tristan Naumann, Hoifung Poon, and Jianfeng Gao. 2023.
\newblock Llava-med: Training a large language-and-vision assistant for biomedicine in one day.
\newblock \emph{Advances in Neural Information Processing Systems}, 36:28541--28564.

\bibitem[{Liu et~al.(2021)Liu, Yin, Wu, Ge, Zhang, and Sun}]{liu2021contrastive}
Fenglin Liu, Changchang Yin, Xian Wu, Shen Ge, Ping Zhang, and Xu~Sun. 2021.
\newblock Contrastive attention for automatic chest x-ray report generation.
\newblock In \emph{Findings of the association for computational linguistics: ACL-IJCNLP 2021}, pages 269--280.

\bibitem[{Lou et~al.(2025)Lou, Yang, Yu, Fu, Han, Huang, and Yu}]{lou2025cxragent}
Jinhui Lou, Yan Yang, Zhou Yu, Zhenqi Fu, Weidong Han, Qingming Huang, and Jun Yu. 2025.
\newblock Cxragent: Director-orchestrated multi-stage reasoning for chest x-ray interpretation.
\newblock \emph{arXiv preprint arXiv:2510.21324}.

\bibitem[{Ma et~al.(2024)Ma, He, Li, Han, You, and Wang}]{medsam}
Jun Ma, Yuting He, Feifei Li, Lin Han, Chenyu You, and Bo~Wang. 2024.
\newblock Segment anything in medical images.
\newblock \emph{Nature Communications}, 15(1):654.

\bibitem[{Moor et~al.(2023)Moor, Huang, Wu, Yasunaga, Dalmia, Leskovec, Zakka, Reis, and Rajpurkar}]{moor2023med}
Michael Moor, Qian Huang, Shirley Wu, Michihiro Yasunaga, Yash Dalmia, Jure Leskovec, Cyril Zakka, Eduardo~Pontes Reis, and Pranav Rajpurkar. 2023.
\newblock Med-flamingo: a multimodal medical few-shot learner.
\newblock In \emph{Machine Learning for Health (ML4H)}, pages 353--367. PMLR.

\bibitem[{Mullappilly et~al.(2024)Mullappilly, Kurpath, Pieri, Alseiari, Cholakkal, Aldahmani, Khan, Anwer, Khan, Baldwin et~al.}]{mullappilly2024bimedix2}
Sahal~Shaji Mullappilly, Mohammed~Irfan Kurpath, Sara Pieri, Saeed~Yahya Alseiari, Shanavas Cholakkal, Khaled Aldahmani, Fahad Khan, Rao Anwer, Salman Khan, Timothy Baldwin, and 1 others. 2024.
\newblock Bimedix2: Bio-medical expert lmm for diverse medical modalities.
\newblock \emph{arXiv preprint arXiv:2412.07769}.

\bibitem[{Najjar(2023)}]{najjar2023redefiningeee}
Reabal Najjar. 2023.
\newblock Redefining radiology: a review of artificial intelligence integration in medical imaging.
\newblock \emph{Diagnostics}, 13(17):2760.

\bibitem[{{OpenAI}(2025)}]{openai2025gpt51}
{OpenAI}. 2025.
\newblock \href {https://openai.com/index/gpt-5-1/} {Introducing {GPT-5.1}}.
\newblock Accessed: 2025-12-27.

\bibitem[{Sellergren et~al.(2025)Sellergren, Kazemzadeh, Jaroensri, Kiraly, Traverse, Kohlberger, Xu, Jamil, Hughes, Lau et~al.}]{sellergren2025medgemma}
Andrew Sellergren, Sahar Kazemzadeh, Tiam Jaroensri, Atilla Kiraly, Madeleine Traverse, Timo Kohlberger, Shawn Xu, Fayaz Jamil, C{\'\i}an Hughes, Charles Lau, and 1 others. 2025.
\newblock Medgemma technical report.
\newblock \emph{arXiv preprint arXiv:2507.05201}.

\bibitem[{Smit et~al.(2020)Smit, Jain, Rajpurkar, Pareek, Ng, and Lungren}]{smit2020chexbertsemb}
Akshay Smit, Saahil Jain, Pranav Rajpurkar, Anuj Pareek, Andrew~Y Ng, and Matthew~P Lungren. 2020.
\newblock Chexbert: combining automatic labelers and expert annotations for accurate radiology report labeling using bert.
\newblock \emph{arXiv preprint arXiv:2004.09167}.

\bibitem[{Tanida et~al.(2023)Tanida, M{\"u}ller, Kaissis, and Rueckert}]{tanida2023interactive}
Tim Tanida, Philip M{\"u}ller, Georgios Kaissis, and Daniel Rueckert. 2023.
\newblock Interactive and explainable region-guided radiology report generation.
\newblock In \emph{Proceedings of the IEEE/CVF Conference on Computer Vision and Pattern Recognition}, pages 7433--7442.

\bibitem[{Wang et~al.(2023)Wang, Liu, Wang, and Zhou}]{wang2023metransformer}
Zhanyu Wang, Lingqiao Liu, Lei Wang, and Luping Zhou. 2023.
\newblock Metransformer: Radiology report generation by transformer with multiple learnable expert tokens.
\newblock In \emph{Proceedings of the IEEE/CVF conference on computer vision and pattern recognition}, pages 11558--11567.

\bibitem[{Wang et~al.(2025)Wang, Wu, Cai, Low, Yang, Li, and Jin}]{wang2025medagent}
Ziyue Wang, Junde Wu, Linghan Cai, Chang~Han Low, Xihong Yang, Qiaxuan Li, and Yueming Jin. 2025.
\newblock Medagent-pro: Towards evidence-based multi-modal medical diagnosis via reasoning agentic workflow.
\newblock \emph{arXiv preprint arXiv:2503.18968}.

\bibitem[{Xu et~al.(2025)Xu, Chan, Li, Aljunied, Yuan, Wang, Xiao, Chen, Liu, Li et~al.}]{xu2025lingshu}
Weiwen Xu, Hou~Pong Chan, Long Li, Mahani Aljunied, Ruifeng Yuan, Jianyu Wang, Chenghao Xiao, Guizhen Chen, Chaoqun Liu, Zhaodonghui Li, and 1 others. 2025.
\newblock Lingshu: A generalist foundation model for unified multimodal medical understanding and reasoning.
\newblock \emph{arXiv preprint arXiv:2506.07044}.

\bibitem[{Yao et~al.(2022)Yao, Zhao, Yu, Du, Shafran, Narasimhan, and Cao}]{yao2022react}
Shunyu Yao, Jeffrey Zhao, Dian Yu, Nan Du, Izhak Shafran, Karthik~R Narasimhan, and Yuan Cao. 2022.
\newblock React: Synergizing reasoning and acting in language models.
\newblock In \emph{The eleventh international conference on learning representations}.

\bibitem[{Yu et~al.(2023)Yu, Endo, Krishnan, Pan, Tsai, Reis, Fonseca, Lee, Abad, Ng et~al.}]{yu2023evaluatingradcliq}
Feiyang Yu, Mark Endo, Rayan Krishnan, Ian Pan, Andy Tsai, Eduardo~Pontes Reis, Eduardo Kaiser Ururahy~Nunes Fonseca, Henrique Min~Ho Lee, Zahra Shakeri~Hossein Abad, Andrew~Y Ng, and 1 others. 2023.
\newblock Evaluating progress in automatic chest x-ray radiology report generation.
\newblock \emph{Patterns}, 4(9).

\bibitem[{Zambrano~Chaves et~al.(2025)Zambrano~Chaves, Huang, Xu, Xu, Usuyama, Zhang, Wang, Xie, Khademi, Yang, Awadalla, Gong, Hu, Yang, Li, Gao, Gu, Wong, Wei, Naumann, Chen, Lungren, Chaudhari, Yeung-Levy, Langlotz, Wang, and Poon}]{ZambranoChaves2025llavarad}
Juan~Manuel Zambrano~Chaves, Shih-Cheng Huang, Yanbo Xu, Hanwen Xu, Naoto Usuyama, Sheng Zhang, Fei Wang, Yujia Xie, Mahmoud Khademi, Ziyi Yang, Hany Awadalla, Julia Gong, Houdong Hu, Jianwei Yang, Chunyuan Li, Jianfeng Gao, Yu~Gu, Cliff Wong, Mu~Wei, and 8 others. 2025.
\newblock \href {https://doi.org/10.1038/s41467-025-58344-x} {A clinically accessible small multimodal radiology model and evaluation metric for chest x-ray findings}.
\newblock \emph{Nature Communications}, 16(1):3108.

\bibitem[{Zhang et~al.(2024{\natexlab{a}})Zhang, Zhou, Yang, Banerjee, Acosta, Miller, Huang, and Rajpurkar}]{zhang2024rexrank}
Xiaoman Zhang, Hong-Yu Zhou, Xiaoli Yang, Oishi Banerjee, Juli{\'a}n~N Acosta, Josh Miller, Ouwen Huang, and Pranav Rajpurkar. 2024{\natexlab{a}}.
\newblock Rexrank: A public leaderboard for ai-powered radiology report generation.
\newblock \emph{arXiv preprint arXiv:2411.15122}.

\bibitem[{Zhang et~al.(2024{\natexlab{b}})Zhang, Zhang, Xie, Li, Dai, Long, Xie, Zhang, Li, and Zhang}]{zhang2024gme}
Xin Zhang, Yanzhao Zhang, Wen Xie, Mingxin Li, Ziqi Dai, Dingkun Long, Pengjun Xie, Meishan Zhang, Wenjie Li, and Min Zhang. 2024{\natexlab{b}}.
\newblock Gme: Improving universal multimodal retrieval by multimodal llms.
\newblock \emph{arXiv preprint arXiv:2412.16855}.

\bibitem[{Zhao et~al.(2024)Zhao, Wu, Zhang, Zhang, Wang, and Xie}]{zhao2024ratescore}
Weike Zhao, Chaoyi Wu, Xiaoman Zhang, Ya~Zhang, Yanfeng Wang, and Weidi Xie. 2024.
\newblock Ratescore: A metric for radiology report generation.
\newblock \emph{arXiv preprint arXiv:2406.16845}.

\bibitem[{Zhu et~al.(2025)Zhu, Wang, Chen, Liu, Ye, Gu, Tian, Duan, Su, Shao et~al.}]{zhu2025internvl3}
Jinguo Zhu, Weiyun Wang, Zhe Chen, Zhaoyang Liu, Shenglong Ye, Lixin Gu, Hao Tian, Yuchen Duan, Weijie Su, Jie Shao, and 1 others. 2025.
\newblock Internvl3: Exploring advanced training and test-time recipes for open-source multimodal models.
\newblock \emph{arXiv preprint arXiv:2504.10479}.

\end{thebibliography}

\end{document}